\documentclass[conference]{IEEEtran}
\IEEEoverridecommandlockouts

\usepackage{amsmath}
\usepackage{algorithm}
\usepackage{algpseudocode}
\usepackage{graphicx}
\usepackage{float}
\usepackage{orcidlink}
\usepackage[font=footnotesize]{caption}
\usepackage{multirow}
\usepackage{booktabs}

\begin{document}

\title{Robust Cooperative Localization in Featureless Environments: A Comparative Study of DCL, StCL, CCL, CI, and Standard-CL}

\author{
\IEEEauthorblockN{
Nivand Khosravi\orcidlink{0009-0004-3749-7249}\IEEEauthorrefmark{1},
Rodrigo Ventura\orcidlink{0000-0002-5655-9562}\IEEEauthorrefmark{1},
Meysam Basiri\orcidlink{0000-0002-8456-6284}\IEEEauthorrefmark{1}
}
\IEEEauthorblockA{\IEEEauthorrefmark{1}
Department of Electrical and Computer Engineering\\
Institute for Systems and Robotics (ISR)\\
Instituto Superior T\'ecnico\\
Lisbon, Portugal\\
\{nivand.khosravi, rodrigo.ventura, meysam.basiri\}@tecnico.ulisboa.pt
}
}

\maketitle
\IEEEpubidadjcol
\begingroup
\renewcommand\thefootnote{}
\footnotetext{\copyright~2026 IEEE. Personal use of this material is permitted. Permission from IEEE must be obtained for all other uses, in any current or future media, including reprinting/republishing this material for advertising or promotional purposes, creating new collective works, for resale or redistribution to servers or lists, or reuse of any copyrighted component of this work in other works.}
\endgroup

\begin{abstract}
Cooperative localization (CL) enables accurate position estimation in multi-robot systems operating in GPS-denied environments. This paper presents a comparative study of five CL approaches: Centralized Cooperative Localization (CCL), Decentralized Cooperative Localization (DCL), Sequential Cooperative Localization (StCL), Covariance Intersection (CI), and Standard Cooperative Localization (Standard-CL). All methods are implemented in ROS and evaluated through Monte Carlo simulations under two conditions: weak data association and robust detection. Our analysis reveals fundamental trade-offs among the methods. StCL and Standard-CL achieve the lowest position errors but exhibit severe filter inconsistency, making them unsuitable for safety-critical applications. DCL demonstrates remarkable stability under challenging conditions due to its measurement stride mechanism, which provides implicit regularization against outliers. CI emerges as the most balanced approach, achieving near-optimal consistency while maintaining competitive accuracy. CCL provides theoretically optimal estimation but shows sensitivity to measurement outliers. These findings provide practical guidance for selecting CL algorithms based on application requirements.
\end{abstract}


\section{Introduction}
Accurate localization is fundamental to the effective operation of autonomous robots. In multi-robot systems, where robots must coordinate to accomplish tasks such as environmental monitoring, search and rescue, and surveillance, localization precision becomes even more critical, as errors in one robot’s estimate can propagate through the network and compromise mission success~\cite{b1}. Traditional localization methods often rely on external reference systems such as GPS, but their effectiveness is limited by signal availability and reliability, particularly in obstructed or indoor environments. Internal localization methods using onboard sensors such as LiDAR and cameras can map surroundings and track robot motion, yet they struggle in sparse, featureless, or dynamic environments.

Cooperative Localization addresses these limitations by enabling robots to share relative observations and jointly estimate their positions, making it particularly effective in featureless or cluttered settings. However, CL introduces additional challenges, including increased computational complexity, reliance on inter-robot communication, and the need to correctly handle cross-correlations between robot estimates~\cite{b2}.

A fundamental challenge in cooperative localization is the treatment of cross-correlations arising from repeated information exchange. When robot $A$ incorporates information from robot $B$, and robot $B$ later uses information derived from robot $A$, their estimates become correlated. If these correlations are not properly accounted for, the filter becomes overconfident and inconsistent, a phenomenon known as data incest or rumor propagation~\cite{b3}. Different CL methods address this issue with varying levels of rigor and computational cost.

In this paper, we present a comparative study of five cooperative localization methods representing distinct strategies for handling cross-correlations:
\begin{enumerate}
    \item \textbf{Centralized CL (CCL)}: Observability-based consistent EKF maintaining full joint state and cross-covariances~\cite{b5}
    \item \textbf{Decentralized CL (DCL)}: Recursive decentralized approach with measurement stride for efficiency~\cite{b4}
    \item \textbf{Sequential CL (StCL)}: Sequential EKF updates incorporating other robots’ uncertainty as additional measurement noise
    \item \textbf{Covariance Intersection CL (CI)}: Conservative fusion guaranteeing consistency under unknown correlations~\cite{b6}
    \item \textbf{Standard CL (Standard-CL)}: Joint EKF assuming independence between robot estimates
\end{enumerate}

All methods are implemented and evaluated in Gazebo-based Monte Carlo simulations with two robots sharing range-bearing measurements under two conditions: weak data association with outliers and robust detection with outlier rejection.

\section{Related Work}

The problem of cooperative localization in multi-robot systems has been extensively studied over the past two decades. Howard et al.~\cite{b1} proposed localization for mobile robot teams using maximum likelihood estimation, establishing foundational approaches for multi-robot cooperative localization.

A fundamental challenge in cooperative localization is handling the cross-correlations that arise when robots share information. Hao et al.~\cite{b3} analyzed this problem and proposed consistent batch fusion for decentralized multi-robot cooperative localization to address filter inconsistency. This has motivated much of the subsequent research in developing methods that either track correlations explicitly or handle them conservatively.

One widely adopted technique to address cross-correlation challenges is Covariance Intersection (CI), introduced by Julier and Uhlmann~\cite{b7}. CI fuses estimates by assuming maximum possible correlation between sources, thereby guaranteeing consistency regardless of the true correlation structure. Carrillo-Arce et al.~\cite{b8} applied CI to decentralized multi-robot cooperative localization, demonstrating its effectiveness in avoiding filter divergence. Kia et al.~\cite{b6} developed a server-assisted distributed cooperative localization framework that leverages covariance intersection for robust estimation over unreliable communication links. Chang et al.~\cite{b9} further investigated resilient and consistent multi-robot cooperative localization with covariance intersection, demonstrating its effectiveness in maintaining filter consistency.

Several approaches have been proposed to mitigate the conservatism of CI while maintaining consistency guarantees. Sijs and Lazar~\cite{b10} introduced Ellipsoidal Intersection (EI), which provides tighter bounds when partial correlation information is available. Li et al.~\cite{b11} proposed Split Covariance Intersection Filter and demonstrated its application to vehicle localization. Noack et al.~\cite{b12} developed Inverse Covariance Intersection (ICI) as an alternative formulation for decentralized data fusion.

Decentralized approaches have received significant attention due to their scalability and robustness to communication failures. Luft et al.~\cite{b4} proposed recursive decentralized localization for multi-robot systems with asynchronous pairwise communication, which forms the basis of the DCL method evaluated in this study. Nerurkar and Roumeliotis~\cite{b13} developed asynchronous multi-centralized cooperative localization to address communication challenges in distributed systems.

Centralized approaches to cooperative localization have been studied extensively due to their theoretical optimality. Huang et al.~\cite{b14} analyzed and improved the consistency of extended Kalman filter based SLAM. Their subsequent work~\cite{b5} developed observability-based consistent EKF estimators for multi-robot cooperative localization, which provides the theoretical foundation for the CCL method in this study. Hao et al.~\cite{b15} further developed graph-based observability analysis for mutual localization in multi-robot systems.

Additional work by Hesch et al.~\cite{b16} on camera-IMU-based localization provided important insights on observability analysis and consistency improvement. Meyer et al.~\cite{b17} investigated cooperative simultaneous localization and synchronization in mobile agent networks, addressing timing challenges in distributed systems.

This study builds upon these foundational works by implementing and systematically comparing five representative CL methods under identical experimental conditions, with particular emphasis on understanding why certain methods exhibit robustness to weak data association.
\section{Models and Problem Formulation}
\label{sec:models}

\subsection{Problem Statement}
Consider a team of $N$ robots moving on a plane. The pose of robot $i$ at discrete time $k$ is
\begin{equation}
\mathbf{X}_{i,k} = [x_{i,k},\, y_{i,k},\, \theta_{i,k}]^\top ,
\end{equation}
where $(x_{i,k}, y_{i,k})$ is position in a global frame and $\theta_{i,k}$ is heading. Each robot maintains an estimate $\hat{\mathbf{X}}_{i,k}$ with covariance $\mathbf{P}_{i,k}$. The objective of cooperative localization is to estimate the joint state $\mathbf{X}_k = [\mathbf{X}_{1,k}^\top,\ldots,\mathbf{X}_{N,k}^\top]^\top$ by fusing odometry and inter-robot measurements, while controlling communication and computation. The key technical challenge is the treatment of cross-correlations $\mathbf{P}_{ij}$ induced by repeated information exchange.

\subsection{Motion Model}
Each robot follows a discrete-time unicycle model:
\begin{equation}
\mathbf{X}_{i,k+1} = \mathbf{X}_{i,k} +
\begin{bmatrix}
V_{i,k}\Delta t\cos\theta_{i,k}\\
V_{i,k}\Delta t\sin\theta_{i,k}\\
\omega_{i,k}\Delta t
\end{bmatrix}
+ \mathbf{w}_{i,k},
\end{equation}
where $\mathbf{u}_{i,k}=[V_{i,k},\,\omega_{i,k}]^\top$ and $\mathbf{w}_{i,k}\sim\mathcal{N}(\mathbf{0},\mathbf{Q}_i)$ with $\mathbf{Q}_i=\mathrm{diag}(\sigma_V^2,\sigma_\omega^2)$. The EKF prediction propagates covariance using the standard Jacobians $\mathbf{F}_i=\partial f/\partial \mathbf{X}_i$ and $\mathbf{G}_i=\partial f/\partial \mathbf{u}_i$.

\subsection{Relative Measurement Model}
When robot $i$ observes robot $j$, it obtains a range-bearing measurement
\begin{equation}
\mathbf{z}_{ij,k} =
\begin{bmatrix}
\rho_{ij,k}\\
\beta_{ij,k}
\end{bmatrix}
+ \mathbf{v}_{ij,k},\qquad
\mathbf{v}_{ij,k}\sim \mathcal{N}(\mathbf{0},\mathbf{R}_{ij}),
\end{equation}
with
\begin{equation}
\rho_{ij,k}=\sqrt{d_x^2+d_y^2},\qquad
\beta_{ij,k}=\mathrm{atan2}(d_y,d_x)-\theta_{i,k},
\end{equation}
where $d_x=x_{j,k}-x_{i,k}$ and $d_y=y_{j,k}-y_{i,k}$, and $\mathbf{R}_{ij}=\mathrm{diag}(\sigma_\rho^2,\sigma_\beta^2)$. The measurement depends on both robot states, with Jacobians $\mathbf{H}_i=\partial h/\partial \mathbf{X}_i$ and $\mathbf{H}_j=\partial h/\partial \mathbf{X}_j$. Note that $\theta_j$ is not directly observed in a single relative range-bearing measurement.

\section{Cooperative Localization Methods}
\label{sec:methods}

All methods share the models in Section~\ref{sec:models} and differ primarily in how they handle cross-correlations.

\subsection{Centralized Cooperative Localization (CCL)}
CCL follows the observability-based consistent EKF framework of Huang et al.~\cite{b5}. A central processor maintains the joint state and full covariance; all odometry and relative measurements are sent to this unit. During prediction, each robot is propagated independently and the cross-covariance evolves as
\begin{equation}
\mathbf{P}_{12}(k+1|k) = \mathbf{F}_1 \mathbf{P}_{12}(k|k) \mathbf{F}_2^\top .
\end{equation}
For a measurement from robot $i$ to robot $j$, the joint Jacobian is
\begin{equation}
\mathbf{H} = \begin{bmatrix} \mathbf{H}_i & \mathbf{H}_j \end{bmatrix}.
\end{equation}
The standard EKF update is then applied to the joint state:
\begin{align}
\mathbf{S} &= \mathbf{H} \mathbf{P}(k+1|k) \mathbf{H}^\top + \mathbf{R}_{ij}, \\
\mathbf{K} &= \mathbf{P}(k+1|k) \mathbf{H}^\top \mathbf{S}^{-1}, \\
\hat{\mathbf{X}}^+ &= \hat{\mathbf{X}} + \mathbf{K} \big(\mathbf{z}_{ij} - h(\hat{\mathbf{X}}_i, \hat{\mathbf{X}}_j)\big), \\
\mathbf{P}^+ &= (\mathbf{I} - \mathbf{K} \mathbf{H}) \mathbf{P}(k+1|k).
\end{align}
CCL is theoretically optimal (minimum-variance for linear Gaussian systems) since it preserves cross-correlations, but it requires centralized communication and introduces a single point of failure.

\subsection{Decentralized Cooperative Localization (DCL)}
DCL follows the recursive decentralized framework of Luft et al.~\cite{b4}. Our implementation uses a \textbf{measurement stride} (stride $=3$), i.e., it processes only every third relative measurement to reduce communication and computation. When robot $i$ observes robot $j$, robot $i$ inflates the measurement covariance to account for its own uncertainty:
\begin{equation}
\mathbf{R}_{eff} = \mathbf{H}_i \mathbf{P}_{ii} \mathbf{H}_i^\top + \mathbf{R}_{ij}.
\end{equation}
Robot $j$ then performs a local update by treating the observation as a measurement with covariance $\mathbf{R}_{eff}$:
\begin{align}
\mathbf{S}_j &= \mathbf{H}_j \mathbf{P}_{jj} \mathbf{H}_j^\top + \mathbf{R}_{eff}, \\
\mathbf{K}_j &= \mathbf{P}_{jj} \mathbf{H}_j^\top \mathbf{S}_j^{-1}.
\end{align}
Beyond efficiency, the stride acts as implicit regularization against outliers and association errors (Section~\ref{sec:discussion}).

\subsection{Sequential Cooperative Localization (StCL)}
StCL performs sequential EKF updates (one robot after the other). The other robot's uncertainty is absorbed into an effective measurement noise:
\begin{equation}
\mathbf{R}_{\mathrm{eff}} = \mathbf{H}_j \mathbf{P}_{jj}\mathbf{H}_j^\top + \mathbf{R}_{ij}.
\label{eq:reff_stcl}
\end{equation}
Cross-covariances are discarded, which can lead to overconfident uncertainty.

\subsection{Covariance Intersection (CI)}
\label{subsec:ci}
CI guarantees consistency when cross-correlations are unknown. Upon receiving an additional estimate $(\hat{\mathbf{x}}^{(2)},\mathbf{P}^{(2)})$ for a robot state, CI fuses it with the local estimate $(\hat{\mathbf{x}}^{(1)},\mathbf{P}^{(1)})$ as
\begin{align}
(\mathbf{P}^{+})^{-1} &= \omega(\mathbf{P}^{(1)})^{-1} + (1-\omega)(\mathbf{P}^{(2)})^{-1}, \\
\hat{\mathbf{x}}^{+} &= \mathbf{P}^{+}\!\left[\omega(\mathbf{P}^{(1)})^{-1}\hat{\mathbf{x}}^{(1)} + (1-\omega)(\mathbf{P}^{(2)})^{-1}\hat{\mathbf{x}}^{(2)}\right],
\end{align}
where $\omega\in[0,1]$ is typically chosen to minimize $\mathrm{tr}(\mathbf{P}^{+})$. CI remains consistent regardless of the true (unknown) correlation structure, at the cost of conservative covariance.

\subsection{Standard Cooperative Localization (Standard-CL)}
Standard-CL applies a joint EKF update while assuming independence ($\mathbf{P}_{12}=\mathbf{0}$). Unlike StCL, it uses the original measurement noise $\mathbf{R}_{ij}$ without inflating it using the other robot's uncertainty, which yields more aggressive covariance reduction and can exacerbate overconfidence.

\subsection{Summary of Differences}
Table~\ref{tab:method_summary} summarizes the five methods. The main distinction is whether cross-correlations are tracked (CCL/DCL), handled conservatively (CI), or ignored (StCL/Standard-CL).
\begin{table}[t]
\centering
\caption{Summary of Cooperative Localization Methods}
\label{tab:method_summary}
\vspace{-4pt}
\begin{tabular}{lcccc}
\toprule
\textbf{Method} & \textbf{Update} & \textbf{$\mathbf{P}_{12}$} & \textbf{Consistent} & \textbf{Complexity} \\
\midrule
CCL & Joint & Tracked & Yes & $O(N^3)$ \\
DCL & Joint + stride & Tracked & Yes & $O(N^3/s)$ \\
StCL & Sequential & Discarded & No & $O(N^2)$ \\
CI & Fusion & Conservative & Yes & $O(N^2)$ + opt. \\
Standard-CL & Joint & Discarded & No & $O(N^2)$ \\
\bottomrule
\end{tabular}
\vspace{-6pt}
\end{table}

\section{Simulation Setup and Metrics}
\subsection{Simulation Setup}
The five cooperative localization methods were implemented in ROS and evaluated through Gazebo-based Monte Carlo simulations with two differential-drive robots. Robot~1 is equipped with a 2D laser scanner (RPLIDAR A2, 6~m range, $360^\circ$ field of view) and wheel encoders, while Robot~2 uses wheel encoders only, reflecting heterogeneous sensing capabilities.

Ground truth trajectories were generated by gazebo model states. For dead-reckoning and EKF prediction, Gaussian noise was injected into the control inputs with $\sigma_v = 0.10$~m/s and $\sigma_\omega = 6^\circ$/s, and the process noise covariance was defined consistently in the control space. Relative measurements were corrupted by isotropic Gaussian noise with $\sigma_z = 0.08$~m. Robots followed curved trajectories with varying linear and angular velocities to ensure sufficient observability. Two measurement conditions were considered, each evaluated over five independent runs:
\begin{itemize}
    \item \textbf{Weak Data Association:} Simple laser-based detection susceptible to outliers.
    \item \textbf{Robust Detection:} Detection with explicit outlier rejection.
\end{itemize}

\subsection{Evaluation Metrics}

\textbf{Position RMSE:} Root Mean Square Error with respect to ground truth, averaged across robots.

\textbf{NEES:} Filter consistency metric defined as
\begin{equation}
\text{NEES} = (\mathbf{x}_{true} - \hat{\mathbf{x}})^\top \mathbf{P}^{-1} (\mathbf{x}_{true} - \hat{\mathbf{x}}),
\end{equation}
with expected value 3 for $n=3$ and 95\% bounds $[0.35,\,9.35]$.

\textbf{NIS:} Measurement consistency metric with expected value 2 for $m=2$.

\section{Results}
\subsection{Performance Comparison}

\begin{figure*}[t]
\centering
\includegraphics[width=0.80\textwidth]{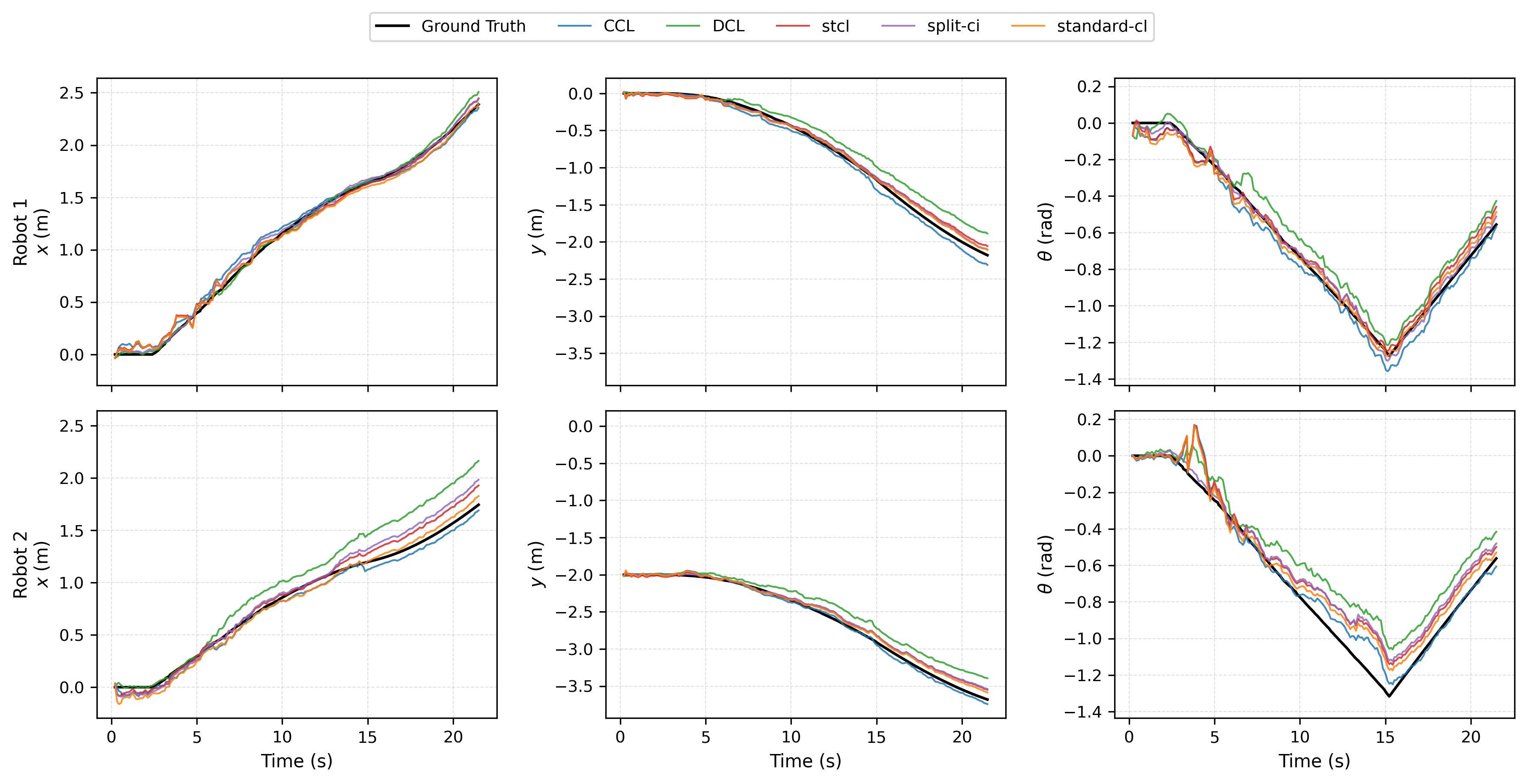}
\caption{State estimation trajectories for a representative run. Top row: Robot 1. Bottom row: Robot 2. DCL (green) shows visible drift due to measurement stride, while other methods track ground truth closely.}
\label{fig:trajectories}
\end{figure*}

Table~\ref{tab:comprehensive} presents the complete performance comparison across all ten experimental trials, while Fig.~\ref{fig:trajectories} illustrates state estimation trajectories for a representative run.

Both robots execute curved paths and among the five methods, DCL exhibits visible position drift due to its measurement stride mechanism, which processes only every third measurement and allows odometry error to accumulate between updates. The remaining methods StCL, Standard-CL, CI, and CCL maintain tighter tracking through more frequent measurement incorporation. Robot 2 consistently shows larger errors across all methods, as it lacks its own sensor and relies entirely on observations from Robot 1.

These trajectory characteristics correspond directly to the quantitative results in Table~\ref{tab:comprehensive}: methods with aggressive measurement updates achieve lower RMSE, while DCL's conservative update strategy yields superior stability under weak data association conditions.

\begin{table*}[t]
\centering
\caption{Performance Comparison Under Robust Detection and Weak Data Association (5 Runs Each)}
\label{tab:comprehensive}
\begin{tabular}{llccccccc}
\toprule
\textbf{Condition} & \textbf{Algorithm} & \textbf{RMSE (m)} & \textbf{Std (m)} & \textbf{Robot 1} & \textbf{Robot 2} & \textbf{Time (ms)} & \textbf{NEES} & \textbf{NIS} \\
\midrule
\multirow{5}{*}{Robust Detection}
 & StCL        & \textbf{0.140} & 0.045 & \textbf{0.133} & \textbf{0.147} & 1.58  & 13.97 & 3.31 \\
 & Standard-CL & 0.148 & 0.062 & \textbf{0.133} & 0.163 & 2.28  & 15.39 & 3.30 \\
 & CI          & 0.183 & 0.065 & 0.179 & 0.187 & 13.94 & \textbf{1.99} & \textbf{0.88} \\
 & CCL         & 0.220 & \textbf{0.044} & 0.163 & 0.277 & 3.05  & 0.70 & 3.59 \\
 & DCL         & 0.223 & 0.099 & 0.181 & 0.266 & \textbf{0.70} & \textbf{0.69} & 3.47 \\
\midrule
\multirow{5}{*}{Weak Association}
 & StCL        & \textbf{0.160} & 0.075 & \textbf{0.157} & \textbf{0.163} & 1.51  & 36.10 & 2.53 \\
 & Standard-CL & 0.174 & 0.076 & 0.177 & 0.170 & 2.29  & 41.83 & 2.53 \\
 & CI          & 0.265 & 0.165 & 0.266 & 0.263 & 13.52 & 5.17 & \textbf{0.71} \\
 & DCL         & 0.282 & \textbf{0.029} & 0.249 & 0.314 & \textbf{0.81} & \textbf{0.89} & 2.52 \\
 & CCL         & 0.335 & 0.194 & 0.338 & 0.331 & 2.94  & 1.14 & 2.70 \\
\bottomrule
\end{tabular}
\end{table*}

\subsection{Position Accuracy and Filter Consistency}

\begin{figure}[t]
\centering
\includegraphics[width=0.65\columnwidth]{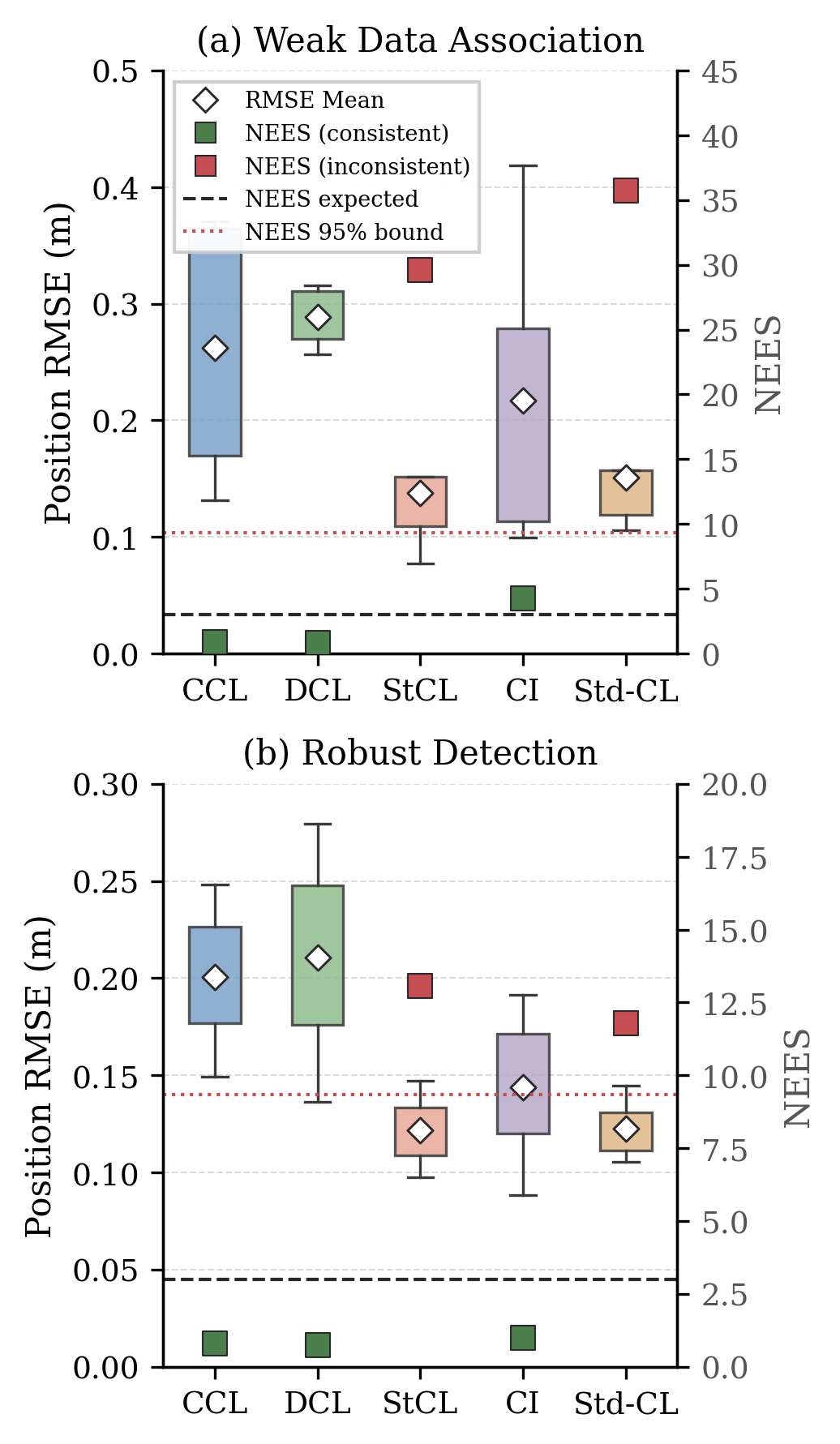}
\caption{RMSE (boxplots, left) and NEES (squares, right) under (a) weak association and (b) robust detection.}
\label{fig:rmse_nees}
\end{figure}

Fig.~\ref{fig:rmse_nees} presents position RMSE distributions (boxplots, left axis) and NEES values (squares, right axis) under both experimental conditions. The dual-axis visualization reveals a clear trade-off between accuracy and consistency across the five methods.

\textbf{Accuracy observations:} StCL and Standard-CL achieve the lowest RMSE (0.12--0.17~m) by aggressively incorporating measurements without tracking cross-correlations. DCL exhibits the lowest RMSE variance under weak association ($\sigma = 0.029$~m), demonstrating exceptional stability against outliers. In contrast, CCL shows high sensitivity to measurement quality, with variance reaching $\sigma = 0.194$~m under weak association as outliers propagate through cross-covariances.

\textbf{Consistency observations:} The right axis reveals a critical distinction among methods. Green squares (CCL, DCL, CI) remain near or below the expected NEES value of 3, indicating proper uncertainty quantification. Notably, CI achieves near-optimal consistency under robust detection (NEES $\approx$ 1.0). Red squares (StCL, Standard-CL) far exceed the 95\% bound, with NEES values of 30--36 under weak association meaning these filters report uncertainty of $\pm$5~cm when actual error is $\pm$15--20~cm.

\textbf{Effect of robust detection:} Improved measurement quality benefits all methods. RMSE reductions range from 12.5\% (StCL) to 34.3\% (CCL), with CCL also achieving 77\% variance reduction. NEES values improve by 61--63\% for the inconsistent methods, though StCL and Standard-CL remain above the consistency bound even with clean measurements.

\textbf{Key finding:} The figure clearly illustrates the accuracy-consistency paradox: methods achieving the lowest RMSE (StCL, Standard-CL) exhibit the worst consistency, while properly consistent methods (CCL, DCL, CI) show higher but more reliable position errors.

\subsection{Computational Cost}

As shown in Table~\ref{tab:comprehensive}, DCL achieves the lowest cost due to its measurement stride mechanism. StCL and Standard-CL offer moderate efficiency, while CI incurs the highest overhead due to fusion weight optimization. All methods meet real-time constraints at 10~Hz.

\section{Discussion}
\label{sec:discussion}

\subsection{The Accuracy-Consistency Trade-off}

StCL and Standard-CL achieve the best accuracy but worst consistency, a paradox arising from ignoring cross-correlations. By treating each measurement as independent, these methods prevent error propagation between robots (improving accuracy) but shrink covariance faster than justified (degrading consistency).

This overconfidence poses safety risks, robots may proceed confidently when position uncertainty is actually high. Any downstream algorithm relying on covariance path planning, collision avoidance, sensor fusion will make decisions based on incorrect uncertainty information.

\subsection{StCL versus Standard-CL}

Both methods discard cross-correlations but differ in their update mechanism. StCL updates robots sequentially using effective measurement noise $\mathbf{R}_{eff} = \mathbf{H}_j \mathbf{P}_j \mathbf{H}_j^\top + \mathbf{R}$, which accounts for the other robot's uncertainty. Standard-CL updates both simultaneously using only $\mathbf{R}$.

Since $\mathbf{R}_{eff} > \mathbf{R}$, StCL produces smaller Kalman gains and slower covariance shrinkage, explaining its slightly better accuracy (0.140~m vs 0.148~m) and consistency (NEES 13.97 vs 15.39). This suggests that even partial uncertainty accounting provides measurable benefits.

\subsection{DCL's Robustness Mechanism}

DCL achieves remarkable stability through its \textbf{measurement stride} mechanism, which processes only every $N$-th measurement (stride = 3 in our implementation). When a measurement arrives, DCL checks a counter: if not divisible by the stride, the measurement is skipped.

Originally designed to reduce communication and computation, measurement stride provides additional robustness benefits:
\begin{itemize}
    \item \textbf{Implicit outlier rejection:} Processing 33\% of measurements means exposure to only one-third of randomly occurring outliers.
    \item \textbf{Temporal decorrelation:} Skipping measurements breaks correlations between consecutive errors from persistent data association failures.
    \item \textbf{Conservative covariance:} Fewer updates result in slower covariance shrinkage, maintaining larger uncertainty bounds that accommodate outliers.
\end{itemize}

This finding suggests measurement stride should be considered a robustness feature, not merely a computational optimization.

\subsection{CI as the Balanced Choice}

CI achieves near-optimal consistency (NEES = 1.99) with competitive accuracy (0.183~m), representing a 31\% penalty compared to StCL. This penalty is the cost of guaranteed consistency regardless of true correlations.

For applications requiring reliable uncertainty path planning through narrow passages, collision avoidance, multi-sensor fusion CI's conservatism ensures reported covariance always bounds true error. The computational overhead is acceptable when consistency guarantees outweigh efficiency concerns.

\subsection{CCL: Optimal but Fragile}

CCL achieves excellent consistency (NEES = 0.70--1.14) by properly tracking correlations, but exhibits fragility under weak association ($\sigma = 0.194$~m). Outliers propagate through $\mathbf{P}_{12}$ to corrupt both estimates. With robust detection, stability improves dramatically ($\sigma \rightarrow 0.044$~m), confirming fragility stems from measurement quality rather than algorithmic limitations.

\subsection{Practical Recommendations}

\textbf{Safety-critical applications:} CI is preferred due to its guaranteed consistency.

\textbf{Unreliable measurements:} DCL offers robust and stable performance; a larger stride can further improve robustness.

\textbf{Reliable sensing conditions:} CCL provides near-optimal estimation accuracy.

\textbf{Limited computational resources:} DCL achieves the best efficiency while remaining consistent.

\textbf{Point-estimate-only use:} StCL or Standard-CL may suffice.
\section{Conclusion}
This paper presented a comparative evaluation of five cooperative localization methods CCL, DCL, StCL, CI, and Standard-CL through extensive Monte Carlo simulations under varying measurement quality conditions. Our analysis revealed several key findings with practical implications:

\textbf{Accuracy-consistency trade-off:} Methods ignoring cross-correlations (StCL, Standard-CL) achieve superior position accuracy but suffer from severe filter inconsistency. This overconfidence poses significant risks for autonomous systems that rely on uncertainty estimates for decision-making.

\textbf{Measurement stride as robustness mechanism:} DCL's practice of processing only a subset of measurements provides implicit regularization against outliers, resulting in exceptional stability under weak data association. This finding suggests that measurement stride should be considered a robustness feature rather than merely a computational optimization.

\textbf{CI as balanced solution:} Covariance Intersection emerged as the most balanced approach, offering guaranteed consistency regardless of unknown correlations while maintaining competitive accuracy. This makes CI well-suited for safety-critical applications requiring reliable uncertainty quantification.

\textbf{CCL fragility:} While CCL provides theoretically optimal estimation by properly tracking all cross-correlations, it exhibits sensitivity to measurement outliers that limits practical applicability without robust detection pipelines.

\textbf{StCL versus Standard-CL:} The sequential update strategy of StCL, which accounts for the other robot's uncertainty through effective measurement noise, provides marginally better performance than Standard-CL's simultaneous update approach.

Future work will extend this comparison to larger robot teams, investigate adaptive stride mechanisms, and conduct real-world experimental validation.

\section*{Acknowledgment}
The authors acknowledge the financial support provided by the Aero.Next Project under Grant Nos. C645727867 and 00000066.

\end{document}